\documentclass[runningheads]{llncs}
\usepackage{graphicx}
\usepackage{subfig}
\usepackage{amsmath}
\usepackage{amssymb}
\usepackage{multirow}
\usepackage{algorithm}
\usepackage{algpseudocode}
\begin{document}
\title{Knowledge-enhanced Iterative Instruction Generation and Reasoning for Knowledge Base Question Answering}
\titlerunning{KBIGER}
\author{Haowei Du \inst{1} \and Quzhe Huang \inst{2} \and Chen Zhang \inst{3} \and Dongyan Zhao \inst{4} \thanks{Corresponding author}}
\authorrunning{Du. Author et al.}

\institute{Peking University, China\\
\email{2001213236@stu.pku.edu.cn} \and
Peking University, China\\
\email{huangquzhe@pku.edu.cn} \and
Peking University, China\\
\email{zhangch@pku.edu.cn} \and 
Peking University, China\\
\email{zhaodongyan@pku.edu.cn} 
}

\maketitle              % typeset the header of the contribution
\begin{abstract}
Multi-hop Knowledge Base Question Answering(KBQA) aims to find the answer entity in a knowledge base which is several hops from the topic entity mentioned in the question. Existing Retrieval-based approaches first generate instructions from the question and then use them to guide the multi-hop reasoning on the knowledge graph. As the instructions are fixed during the whole reasoning procedure and the knowledge graph is not considered in instruction generation, the model cannot revise its mistake once it predicts an intermediate entity incorrectly. To handle this, we propose \textbf{KBIGER}(\textbf{K}nowledge \textbf{B}ase \textbf{I}terative Instruction \textbf{GE}nerating and \textbf{R}easoning), a novel and efficient approach to generate the instructions dynamically with the help of reasoning graph. Instead of generating all the instructions before reasoning, we take the $(k-1)$-th reasoning graph into consideration to build the $k$-th instruction. In this way, the model could check the prediction from the graph and generate new instructions to revise the incorrect prediction of intermediate entities. We do experiments on two multi-hop KBQA benchmarks and outperform the existing approaches, becoming the new-state-of-the-art%\red{how much}
. Further experiments show our method does detect the incorrect prediction of intermediate entities and has the ability to revise such errors.

\end{abstract}
\section{Introduction}
%\red{knowledge graph or knowledge base? Instruction Generation or Instruction generating? Be careful and consistent!!!}

%\red{What is the meaning of ``Error Accumulation``? You use this phrase everywhere, but you never explain it.}

%\red{In equations, vectors and matrices should be bold...}

Knowledge Base Question Answering(KBQA) is a challenging task that aims to answer the natural language questions with the knowledge graph. With the fast development of deep learning, researchers leverage end-to-end neural networks \cite{liang2016neural,sun2018open} to solve this task by automatically learning entity and relation representations, followed by predicting the intermediate or answer entity. %\red{Why learning entity and relation representations could solve KBQA?} 
Recently for the KBQA community, there have been more and more interests in solving complicated questions where the answer entities are multiple hops away from the topic entities.
One popular way to solve multi-hop KBQA is the information retrieval-based methods, which generate instructions from questions and then retrieve the answer from the Knowledge Graph by using the instructions to guide the reasoning\cite{he2021improving,shi2021transfernet}. 
% use the instructions to guide the reasoning over knowledge graph for retrieving the answer. 
Although achieving good performance on multi-hop KBQA,
existing retrieval-based methods treat instruction generation and reasoning as two separate components. 
Such methods first use the question solely to generate the instructions for all the hops at once and then use them to guide the reasoning. As the instructions are fixed during the whole reasoning procedure and the knowledge graph is not considered in instruction generation, the model cannot revise its mistake once it predicts an intermediate entity incorrectly. When the model reasons to an incorrect intermediate entity, the fixed instruction will guide the reasoning from the wrong prediction, which induces error accumulation.
% If at one step the model predicts an incorrect intermediate entity, the instruction generated beforehand will guide the reasoning around the incorrect intermediate entity, which brings in error accumulation.

\begin{figure}[t!]
\centering
\includegraphics[width=0.5\textwidth]{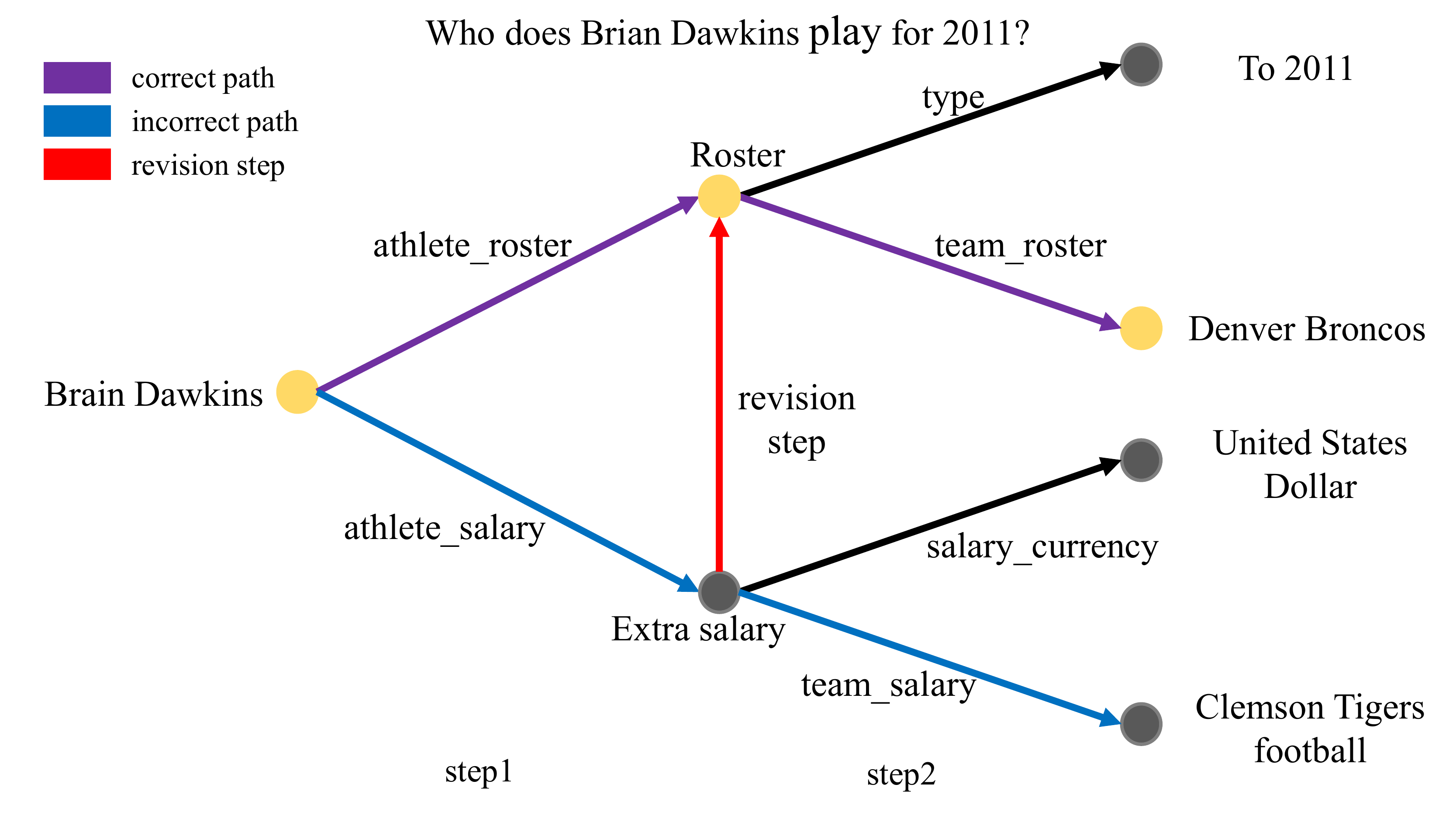}
\caption{An example from WebQSP dataset. The red arrows denote the right reasoning path, the blue arrows denote a wrong reasoning path and the purple arrow denotes the revision in our approach. }%\red{purple?}}
\label{case 1}
\vspace{-2em}
\end{figure}
We take an example in WebQSP, as shown in Figure \ref{case 1}, to show the importance of knowledge graph on predicting intermediate entity and revising error of incorrect predictions. For the query question ``Who does Brian Dawkins play for 2011'', its topic entity is the football player ``Brian Dawkins''. Both ``play'' and ``2011'' can be attended to in the instruction generated at the first step. The baseline method NSM \cite{he2021improving} reasons to a wrong intermediate entity ``extra salary'', perhaps because the model mistakes the time constraint ``2011'' as a regular number and chooses the ``extra\_salary'' entity, which is related with number and currency. At the second step, the instructions by NSM continue to guide the reasoning from the wrong intermediate entity and get the wrong answer ``Clemson Tigers football'', which does not satisfy the time constraint. However, with the knowledge graph information that the ``roster'' entity connects to the type entity ``to 2011'' and the named entity  ``Denver Broncos (football team)''  but the entity ``extra\_salary'' is not linked to another entity by relation “team\_roster", the instructions could revise the error by re-selecting the ``roster'' as the intermediate entity and linking the predicate ```play'' with the relation ``team\_roster'' to derive the answer entity ``Denver Broncos''.

To introduce the knowledge graph into generating instructions from the query question, we propose our approach, \textbf{K}nowledge \textbf{B}ase \textbf{I}terative Instruction \textbf{GE}nerating and \textbf{R}easoning(KBIGER). Our method has two components, the instruction generation component and the reasoning component. At each step, we generate one instruction and reason one hop over the graph under the guidance of the instruction. 
To generate the $k$-th instruction, we take both the question and the $(k-1)$-th reasoning graph into consideration. In this way, our model could obtain results of the last reasoning step and will be able to revise the possible mistakes by generating the new instruction.
% We utilize the last instruction and reasoning graph to attend to part of the query question and generate the new instruction. 
Then we utilize the instruction created
to extend the reasoning path. Besides, we also adopt distilling learning to enhance the supervision signal of intermediate entity, following \cite{he2021improving}.
% For purpose of enhancing supervision signal of intermediate entity distribution, we adopt a teacher-student framework where the student network treats the intermediate entity predicted by the teacher as a soft label. 
We do experiments on two benchmark datasets in the field of KBQA and our approach outperforms the existing methods by 1.0 scores Hits@1 and 1.0 scores F1 in WebQSP, 1.4 score and 1.5 score in CWQ, %by \red{how much and in which dataset}, 
becoming the new state-of-the-art.

Our contributions can be concluded as three folds:
\begin{enumerate}
    \item We are the first to consider the reasoning graph of previous steps when generating the new instruction, which makes the model be able to revise the errors in reasoning.
    \item We create an iterative instruction generation and reasoning framework, instead of treating instruction generation and reasoning as two separate phases as current approaches. This framework could fuse information from question and knowledge graph in a deeper way.
    \item  Our approach outperforms existing methods in two benchmark datasets in this field, becoming the new state-of-the-art.
\end{enumerate}
% \textbf{1.} We are the first to consider the reasoning graph of previous steps when generating the new instruction, which makes the model be able to revise the errors in reasoning.\\
% \textbf{2.} We create an iterative instruction generation and reasoning framework, instead of treating instruction generation and reasoning as two separate phases as current approaches. This framework could fuse information from question and knowledge graph in a deeper way.\\
% \textbf{3.} Our approach outperforms existing methods in two benchmark datasets in this field, becoming the new state-of-the-art.
\section{Related Work}
Over the last decade, various methods have been developed for the KBQA task. Early works utilize machine-learned or hand-crafted modules like entity recognition and relation linking to find out the answer entity \cite{yih2015semantic,berant2013semantic,dong2015question,ferrucci2010building}. With the popularity of neural networks, recent researchers utilize end-to-end neural networks to solve this task. They
can be categorized into two groups: semantic parsing based methods \cite{yih2015semantic,liang2016neural,guo2018dialog,saha2019complex} and information retrieval based methods \cite{zhou2018interpretable,zhao2019simple,cohen2020scalable,qiu2020stepwise}. 
Semantic parsing methods convert natural language questions into logic forms by learning a parser and predicting the query graph step by step. However, the predicted graph is dependent on the prediction of last step and if at one step the model inserts the incorrect intermediate entity into the query graph, the prediction afterward will be unreasonable. Information retrieval-based methods retrieve answers from the knowledge base by learning and comparing representations of the question and the graph. 
\cite{miller2016key} utilize Key-Value Memory Network to encode the questions and facts to retrieve answer entities. 
To decrease the noise in questions, \cite{zhang2018variational} introduces variance reduction into retrieving answers from the knowledge graph. Under the setting of supplemented corpus and incomplete knowledge graph, \cite{sun2019pullnet} proposes PullNet to learn what to retrieve from a corpus. \cite{shi2021transfernet} proposes TransferNet to support both label relations and text relations in a unified framework, For purpose of enhancing the supervision of intermediate entity distribution, \cite{he2021improving} adopts the teacher-student network in multi-hop KBQA. However, it ignores the utility of knowledge graph in generating information from questions and the instructions generated are fixed in the whole reasoning process, which undermine the ability to revise incorrect prediction of intermediate entities.

\section{Preliminary}
In this part, we introduce the concept of knowledge graph and the definition of multi-hop knowledge base question answering task(KBQA). \\
\textbf{Knowledge Graph(KG)}
A knowledge graph contains a series of factual triples and each triple is composed of two entities and one relation. A knowledge graph can be denoted as $G = \{(e,r,e')| e,e' \in E, r\in R\}$, where $G$ denotes the knowledge graph, $E$ denotes the entity set and $R$ denotes the relation set. A triple $(e,r,e')$ means relation $r$ exists between the two entities $e$ and $e'$. We use $N_e$ to denote the entity neighbourhood of entity $e$, which includes all the triples including $e$, $i.e.,N_e = \{(e',r,e) \in G\} \cup \{(e,r,e') \in G\}$. \\
\textbf{Multi-hop KBQA}
%\red{I suggest you rewrite this part.}
Given a natural language question $q$ that is answerable using the knowledge graph, the task aims to find the answer entity. The reasoning path starts from \emph{topic entity}(i.e., entity mentioned in the question) to the answer entity. Other than the topic entity and answer entity, the entities in the reasoning path are called \emph{ intermediate entity}. If two entities are connected by one relation, the transition from one entity to another is called one hop. In multi-hop KBQA, the answer entity is connected to the topic entity by several hops. %The maximum number of hops $n$ between the topic entity and answer is given in each dataset. 

\section{Methodology}
\begin{figure*}[h]
\centering
\includegraphics[width=0.6\textwidth]{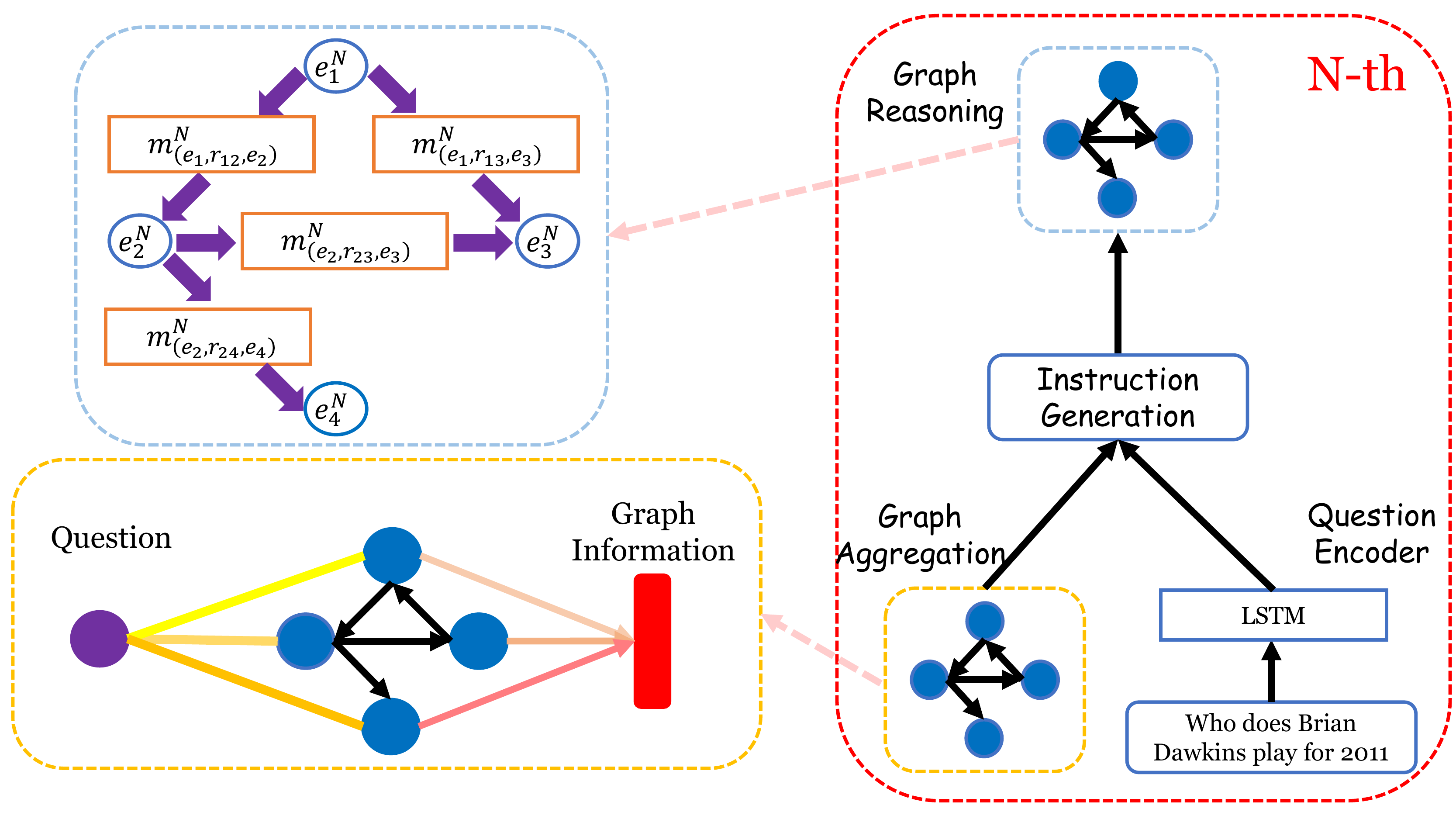}
\caption{Method Overview}
\label{pipeline}
\vspace{-2em}
\end{figure*}

Our approach is made up of two components, Instruction Generation and Graph Reasoning. In the former component, we utilize the query and knowledge graph to generate some instructions for guiding the reasoning. In the graph reasoning component, we adopt GNN to reason over the knowledge graph. We also apply the teacher-student framework, following \cite{he2021improving}, to enhance the intermediate supervision signal.
For each question, a subgraph of the knowledge graph is constructed by reserving the entities that are within $n$ hops away from the topic entity to simplify the reasoning process, where $n$ denotes the maximum hops between the topic entity and the answer.

% Existing approaches treats instruction generation and reasoning as two separate phase and fail to utilize the information of the knowledge graph in generating instructions. The instructions are fixed in the whole reasoning process, which increases the error accumulation of wrong prediction. To solve this problem and enhance the information communication between the query and knowledge graph, we propose our approach KBIGER \ref{pipeline}. Following \cite{he2021improving}, we utilize the teacher-student learning framework to provide (pseudo) supervision signals which are intermediate entity distributions in our task. \subsection{Iterative Instruction Generation and Reasoning}
%\red{I think you should introduce instruction generation first. Entity Initialization should be part of graph reasoning.}

\subsection{Instruction Generation Component}
The objective of this component is to utilize the text of query question and the information of knowledge graph to construct a series of instructions $\{\mathbf{i_k}\}_{k=1}^{n} \in \mathbb{R}^d$, where $\mathbf{i_k}$ denotes the $k$-th instruction for guiding the reasoning over the knowledge graph. 
%where $n$ denotes the number of hops between the topic entity and the answer entity.% 
We use BiLSTM to encode the question to obtain a contextual representation of each word in the question, where the representation of $i$-th word is denoted as $\mathbf{h_i}$. We utilize the last hidden state $\mathbf{h_l}$ as the semantic representation of the question,i.e.,$\mathbf{q} = \mathbf{h_l}$. 
Both question and the aggregation result of the previous reasoning graph are used to generate instructions. The instruction generated is adapted to the reasoning graph instead of fixed in the whole reasoning process. We utilize the attention mechanism to attend to different parts of the query at each step.
We construct instructions as follows:
\begin{align}
  \mathbf{i^{(k)}} &= \sum_{j=1}^l \alpha_j^{(k)} \mathbf{h_j} \\
        \alpha_j^{(k)} &= \mathbf{softmax_j}(\mathbf{W_\alpha}( \mathbf{q^{(k)}} \odot \mathbf{h_j}))\\
    \mathbf{q^{(k)}} &= \mathbf{W^{(k)}}[\mathbf{i^{(k-1)}};\mathbf{q}; \mathbf{e_{graph}^{(k-1)}}]+\mathbf{b^{(k)}} 
    \end{align}
where $\mathbf{e_{graph}^{(k-1)}}\in\mathbb{R}^d$ denotes the representation of $(k-1)$-th reasoning graph which we will explain below and $\mathbf{W^{(k)}} \in \mathbb{R}^{d\times3d}, \mathbf{W_\alpha} \in \mathbb{R}^{d\times d},\mathbf{b^{(k)}} \in \mathbb{R}^d$ are learnable parameters. 
%n this process, the model attends to different parts of the query question at each step. We utilize information from the query question and knowledge graph by attention mechanism to decide the proper relation for the following reasoning step. 

%\red{I think you should move Graph Aggregation here, explain how to get $e_graph^k$}
 \subsection{Graph Aggregation}
In this stage, we combine the query question and KB entity representation to generate the whole graph representation. We adopt the attention mechanism to assign weights to each entity in the reasoning graph and aggregate them into a graph representation. In this way, the model is aware of the graph structure of intermediate entities and has the ability to generate new instructions to revise the incorrect prediction.  
\begin{align}
 \mathbf{ e_{graph}^{(k)}} &= \sum_{e \in E} \alpha_e^{(k)}\mathbf{e^{(k)}} \\
      \alpha_e^{(k)} &= \frac{\mathbf{exp}(\beta_e^{(k)})}{ \sum_{e' \in E} \mathbf{exp}(\beta_{e'}^{(k)})} \\
     \beta_e^{(k)} &= \mathbf{q} \cdot  (\mathbf{W_{gate}}\mathbf{e^{(k)}} + \mathbf{b_q})
    \end{align}
where $\mathbf{W_{gate}} \in \mathbb{R}^{d\times d},\mathbf{b_q} \in \mathbb{R}^d$ are parameters to learn and ``$ \cdot $'' denotes inner product.

\subsection{Entity Initialization}
 We believe the relations involving the entity contain important semantic information of the entity, which can be used to initialize entity.
 We set the initial entity embedding for each entity in the subgraph by considering the relations involving it:
\begin{align}
 \mathbf{e^{(0)}} = \sigma(\sum_{(e',r,e) \in N_{e}} \mathbf{W_{E}} \cdot \mathbf{r}) \label{eq.7}\end{align}
where $(e',r,e)$ denotes a triple in the subgraph, $\mathbf{r} \in \mathbb{R}^d$ denotes the embedding vector of relation $r$ and $\mathbf{W_{E}}$ is parameter to learn. 

\subsection{Reasoning Component}
The objective of this component is to reason over knowledge graph under the guidance of the instruction obtained by the previous instruction generation component. %At the $k$-th step, the input is the instruction derived from the $k$-th instruction component $i^{(k)}$ as well as the distribution and representation of entities from the $(k-1)$-th reasoning component i.e.,${P^{(k-1)}}\ {E^{(k-1)}}$. The output is the distribution of the $k$-th intermediate entity and the updated entity representation i.e.,${P^{(k)}}\ {E^{(k)}}$.
First, for each triple $(e',r,e)$ in the subgraph, we learn a matching vector between the triple and the current instruction $i^{(k)}$:
\begin{equation}
   \mathbf{ m^{(k)}_{(e',r,e)}} = \sigma(\mathbf{i^{(k)}} \odot \mathbf{W_Rr})
\end{equation} where $\mathbf{r}$ denotes the embedding of relation r and $\mathbf{W_R}$ are parameters to learn.
Then for each entity $e$ in the subgraph, we multiply the activating probability of each neighbour entity $e'$ by the matching vector $\mathbf{m^{(k)}_{(e',r,e)}} \in \mathbb{R}^d$ and aggregate them as the representation of information from its neighbourhood:
\begin{align}
  \mathbf{\hat{e}^{(k)}} = \sum_{(e',r,e)} p_{e'}^{(k-1)}\mathbf{m^{(k)}_{(e',r,e)}}
\end{align}
The activating probability of entities is derived from the distribution predicted by the previous reasoning component. We concatenate the previous entity representation with its neighbourhood representation and pass into a MLP network to update the entity representation:
\begin{align}
  \mathbf{ e^{(k)}}  = \mathbf{MLP}(\mathbf{e^{(k-1)}};\mathbf{\hat{e}^{(k)}})
\end{align}
Then we compute the distribution of the $k$-th intermediate entities as follows:
\begin{align}
  \mathbf{p^{(k)}} = \mathbf{softmax}(\mathbf{E^{(k)}} \mathbf{W_E})
\end{align}
where each column of $\mathbf{E^{(k)}}$ is the updated entity embedding $\mathbf{e^{(k)}} \in \mathbb{R}^d$ and $\mathbf{W_E} \in \mathbb{R}^d$ is parameter to learn.

\subsection{Algorithm}
To conclude the above process of iterative instruction generation and reasoning, we organize it as Algorithm 1. We generate the instruction from the question and reason over knowledge graph alternately. The instruction component sends instructions to guide the reasoning and the reasoning component provides the instruction component with knowledge from a related graph. This mechanism allows two components to mutually communicate information with each other.  
\begin{algorithm}[t]
\caption{Iterative instruction generation and reasoning} %算法的名字
\begin{algorithmic}[1]
\State For each entity in subgraph, initialize entity embedding by Eq.\ref{eq.7}. Let $n$ denote the number of hops to reason. Use Glove and LSTM to obtain word embedding ${\mathbf{h_j}}$ and question embedding $\mathbf{q}$ for the query question. 
\For{k = 1,2,...,n}
\State Generate instruction based on $\mathbf{i^{(k-1)}}$ and $\mathbf{E^{(k-1)}}$, get $\mathbf{i^{(k)}}$
\State Reason over knowledge graph based on $\mathbf{i^{(k)}},\mathbf{P^{(k-1)}},\mathbf{E^{(k-1)}}$ and get $\mathbf{P^{(k)}},\mathbf{E^{(k)}}$
\EndFor
\State Based on the final entity distribution $\mathbf{P^{(n)}}$, take out the entity that has the activating probability over given threshold as answer entities.
\end{algorithmic}
\end{algorithm}

\subsection{Teacher-Student Framework}
 We adopt a teacher-student framework in our approach to enhance the supervision of intermediate entity distribution following \cite{he2021improving}. Both the teacher network and student network have the same architecture where the instruction generation component and reasoning component progress iteratively. The teacher network learns the intermediate entity distribution as the supervision signal to guide the student network. 
 %After the teacher network is trained, the intermediate entity distribution of teacher network can be seen as the soft label, so the intermediate entity distribution predicted by the student network should be similar with the soft label from the teacher. 
 The loss function of the student network is designed as follows:
\begin{align}
    L_1 &= \mathbf{D_{KL}}(\mathbf{P^{(n)}_s},\mathbf{P^*}) \\
    L_2  &= \sum_{k=1}^n \mathbf{D_{KL}}(\mathbf{P^{(k)}_s},\mathbf{P^{(k)}_t}) \\
    L_s &= L_1 + \lambda L_2 \label{eq.14}
\end{align}
$\mathbf{D_{KL}(\cdot)}$ denotes the Kullback-Leibler divergence. $\mathbf{P^{(k)}_s}$ and $\mathbf{P^{(k)}_t}$ denotes the predicted distribution of the $k$-th intermediate entity by student network and teacher network. $\mathbf{P^*}$ denotes the golden distribution of answer entity. $\lambda$ is a hyper-parameter to tune.

\section{Experiments}
\begin{table}
\vspace{-2em}
\centering
\begin{tabular}{|l|c|c|c|c|}
\hline
\textbf{Datasets} & \textbf{train} & \textbf{dev} & \textbf{test} & \textbf{entities}\\
\hline
WebQSP & 2,848 & 250 & 1,639 & 1,429.8\\
CWQ &  27,639 & 3,519 & 3,531 & 1,305.8  \\
\hline
\end{tabular}
\caption{Statistics about the datasets. The column ``\textbf{entities}'' denotes the average number of entities in the subgraph for each question}
\label{statistics about the datasets}
\vspace{-5em}
\end{table}
\subsection{Datasets, Evaluation Metrics and Implementation Details}
We evaluate our method on two widely used datasets, WebQuestionsSP and Complex WebQuestion. Table \ref{statistics about the datasets} shows the statistics about the two datasets.

\textbf{WebQuestionsSP(WebQSP)}\cite{yih2015semantic} includes 4,327 natural language questions that are answerable based on Freebase knowledge graph \cite{bollacker2008freebase}, which contains millions of entities and triples. The answer entities of questions in WebQSP are either 1 hop or 2 hops away from the topic entity. Following \cite{saxena2020improving}, we prune the knowledge graph to contain the entities within 2 hops away from the mentioned entity. On average, there are 1,430 entities in each subgraph.

\textbf{Complex WebQuestions(CWQ)}\cite{talmor2018web} is expanded from WebQSP by extending question entities and adding constraints to answers. It has 34,689 natural language questions which are up to 4 hops of reasoning over the graph. Following \cite{sun2019pullnet}, we retrieve a subgraph for each question using PageRank algorithm. On average, there are 1,306 entities in each subgraph. 

Following \cite{sun2018open,sun2019pullnet,he2021improving}, we treat the multi-hop KBQA task as a ranking task. For each question, we select a set of answer entities based on the distribution predicted. We utilize two evaluation metrics Hits@1 and F1 that are widely applied in the previous work \cite{sun2018open,sun2019pullnet,shi2021transfernet,saxena2020improving}. Hits@1 measures the percent of the questions where the predicted answer entity that has maximum probability is in the set of ground-truth answer entities. F1 is computed by use of the set of predicted answer entities and the set of ground-truth answers. Hits@1 focuses on the entity with the maximum probability in the final distribution predicted and F1 focuses on the complete answer set. 

Before training the student network, we pre-train the teacher network on the multi-hop KBQA task. We optimize all models with Adam optimizer, where the batch size is set to 32 and the learning rate is set to 7e-4. The reasoning steps are set to 3 for WebQSP and 4 for CWQ. The coefficient in Eq.\ref{eq.14} is set to 0.05. The hidden size of LSTM and GNN is set to 128.  

\subsection{Baselines to Compare}% \red{or Baselines?}}
\textbf{KV-Mem}\cite{miller2016key} takes advantage of Key-Value Memory Networks to encode knowledge graph triples and retrieve the answer entity.\\
\textbf{GraftNet}\cite{sun2018open} uses a variation of graph convolution network to update the entity embedding and predict the answer.\\
\textbf{PullNet}\cite{sun2019pullnet} improves GraftNet by retrieving relevant documents and extracting more entities to expand the entity graph.\\
\textbf{EmbedKGQA}\cite{saxena2020improving} utilizes pre-trained knowledge embedding to predict answer entities based on the topic entity and the query question. \\
\textbf{NSM}\cite{he2021improving} takes use of the teacher framework to provide the distribution of intermediate entities as supervision signs for the student network. However, NSM fails to consider the utility of the knowledge graph in generating instructions from the question, which constrains the performance.\\
\textbf{TransferNet}\cite{shi2021transfernet} unifies two forms of relations, label form and text form to reason over knowledge graph and predict the distribution of entities.

\subsection{Results}
\begin{table}
\vspace{-1em}
\centering
\begin{tabular}{lcccc}
\hline
\multirow{2}{1cm}{\textbf{Models}} & \multicolumn{2}{c}{\textbf{WebQSP}} & \multicolumn{2}{c}{\textbf{CWQ}}\\
& Hits@1 & F1 & Hits@1 & F1\\
\hline
KV-Mem & 46.7  & 38.6  & 21.1 & - \\
GraftNet & 67.8 & 62.8 & 32.8 & - \\
PullNet & 68.1 & - & 45.9 & - \\
EmbedKGQA  & 66.6 & - & - & - \\
TransferNet & 71.4 & - &  48.6 & -\\
NSM &  74.3 &  67.4 &  48.8  & 44.0\\
\hline
Ours & \textbf{75.3} & \textbf{68.4} & \textbf{50.2} & \textbf{45.5} \\
\hline
\end{tabular}
\caption{\label{results}
Results on two benchmark datasets compared with several competitive methods proposed in recent years. The baseline results are from original papers. Our approach significantly outperforms NSM, where $p$-values of Hits@1 and F1 are 0.01 and 0.0006 on WebQSP, 0.002 and 0.0001 on CWQ.
}
\vspace{-3em}
\end{table}
The results of different approaches are presented in Table~\ref{results}, by which we can observe the following conclusions:
%\paragraph{(1)}Overall, for most approaches, the performance of WebQSP dataset is better than the CWQ dataset because the questions in CWQ needs more hops of reasoning and more constrains are attached. Among the baselines, $NSM+h$ performs best because of the utility of teacher-student framework to provide supervision signal.s for intermediate entities%
Our approach outperforms all the existing methods on both datasets and evaluation metrics, becoming the new state-of-the-art. It is efficient to introduce the information from knowledge graph into generating instructions from the query question and iteratively proceed the instruction generation component with the reasoning component. Our approach outperforms the previous state-of-the-art NSM by $1.0$ Hits@1 score and $1.0$ F1 score in WebQSP as well as  $1.4$ Hits@1 score and $1.5$ F1 score in CWQ. CWQ composes more complicated query questions and our model is good at answering complex questions with more hops of reasoning. Compared to the Hits@1 metric, the F1 metric counts on the prediction of the whole set of answer entities instead of the answer that has the maximum probability. The performance of our model in F1 metric shows it can predict the whole answer set instead of just one answer. 
\section{Analysis}
%\red{It seems that your improvement mainly comes from the teacher-student framework. The results of ablation 2 are worse than TransferNet and NSM in Table 2}

In this part, we do two ablation studies and evaluate the effectiveness of our approach on revising the incorrect prediction of intermediate entities. Furthermore, we give an example from WebQSP to show the revision of incorrect prediction for intermediate entities.
\subsection{Ablation Study and Error Revision}
To demonstrate the effectiveness of different components of our approach, we do the following ablation study on WebQSP and CWQ datasets:\\
\textbf{Ablation study 1}In this model, we revise the instruction computation by removing the component of graph entity and deriving all the $n$ instructions before the reasoning: 
\begin{align}
\mathbf{q^{(k)}}&= \mathbf{W^{(k)}}[\mathbf{i^{(k-1)}};\mathbf{q}]+\mathbf{b^{(k)}}\\
    \alpha_j^{(k)} &= \mathbf{softmax_j}(\mathbf{W_\alpha}(\mathbf{q^{(k)}} \odot \mathbf{h_j}))\\
    \mathbf{i^{(k)}} &= \sum_{j=1}^l \alpha_j^{(k)}\mathbf{h_j}
    \end{align}
where $n$ denotes the number of hops between answer and topic entity. We can see in this model, the instruction generated from questions do not consider the information of the knowledge graph, and the instructions are fixed in the whole reasoning process. Regardless of what the structure of knowledge graph is, the instructions sent by questions are the same. \\
\textbf{Ablation study 2}
In this model, we remove the teacher network and only use the hard label in datasets for training. %The teacher network shares the same model architecture and number of parameters with the student network in our approach.
%\subsection{Results of Ablation study}
\begin{table}
\vspace{-1.5em}
\centering
\begin{tabular}{lcccc}
\hline
\multirow{2}{1cm}{\textbf{Models}} & \multicolumn{2}{c}{\textbf{WebQSP}} & \multicolumn{2}{c}{\textbf{CWQ}}\\
& Hits@1 & F1 & Hits@1 & F1 \\
\hline
-iterative & 72.1  & 66.4  & 47.7 & 43.6 \\
-teacher& 70.4 & 64.8 & 48.2 & 43.3 \\
\hline
KBIGER & \textbf{75.3} & \textbf{68.4} & \textbf{50.2} & \textbf{45.5} \\
\hline
\end{tabular}
\caption{\label{Ablation study} Results of ablation models on WebQSP and CWQ.}
\vspace{-3em}
\end{table}

\begin{figure}[h] 
\vspace{-2.5em}
\centering
\includegraphics[width=0.2\textwidth]{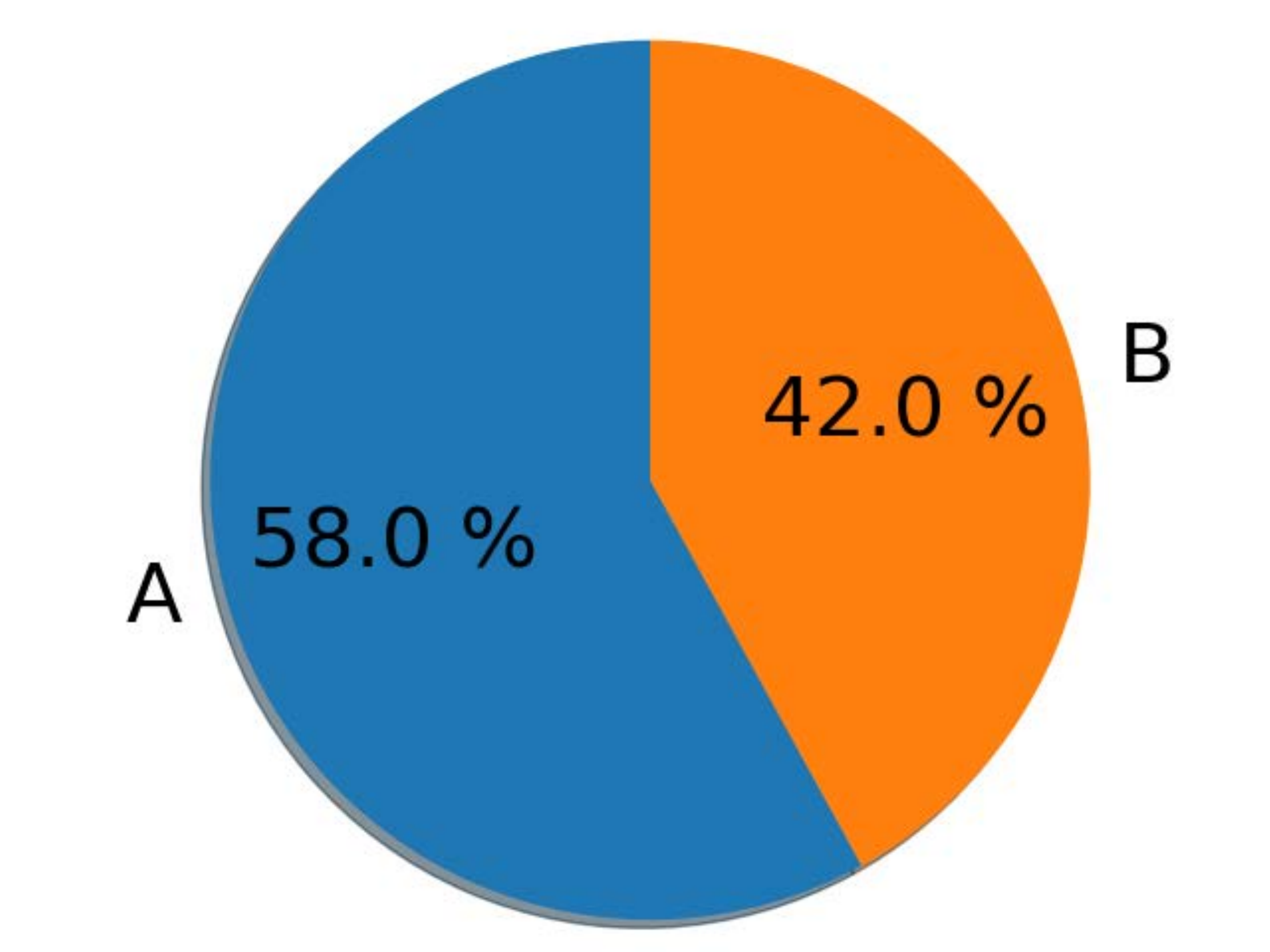}
\caption{The proportion of two groups of cases in WebQSP dataset where we get the correct answer but NSM fails. Group A represents the cases where we revise the error and answer correctly; Group B denotes the cases where we predict all the intermediate entities correctly.}
\label{Percentage of different cases}
\vspace{-2em}
\end{figure}
As shown in table \ref{Ablation study}, we can see if we remove the information of knowledge graph from the instruction generation component, the performance of Hits@1  and F1 on two benchmark datasets will fall down by about 2 points. This reveals that the knowledge graph structure is important for generating flexible instructions from query questions.
Without the student network, the evaluation results are lower than our approach. It demonstrates that the supervision signal of intermediate entity distribution can improve the prediction of intermediate entities and answer entities. Though the lack of the student network, ablation model 2 achieves similar results with ablation model 1, which shows the efficiency of utilizing the information of knowledge graph to generate flexible instructions from questions. 

To explore whether the introduction of knowledge graph into instruction generation could identify the incorrect prediction of intermediate entities and revise them,  we annotated the intermediate entity of the multi-hop data cases where the baseline NSM fails to answer the question correctly but we get the right answer. By checking the reasoning process, we classify the cases into 2 groups and compute the proportion: 
\textbf{1.} the cases where we fail to predict the intermediate entity at first but revise the error in the following steps and answer the question correctly.
\textbf{2.} the cases where we answer the question correctly by predicting all the intermediate entities correctly.

As shown in figure \ref{Percentage of different cases}, among the cases where our model predicts correctly and baseline NSM fails to answer,  58\% are cases where we revise the error of incorrect prediction of an intermediate entity. which shows the efficiency of introducing the knowledge graph into instruction generation in an iterative manner on decreasing the error accumulation along the reasoning path.

\subsection{Case Study}
In this section, we take one case from WebQSP to show the effectiveness of our method in revising the incorrect prediction of intermediate entities.
\begin{figure}[h] 
\vspace{-2em}
\begin{minipage}[h]{0.5\linewidth}
\includegraphics[width=\linewidth]{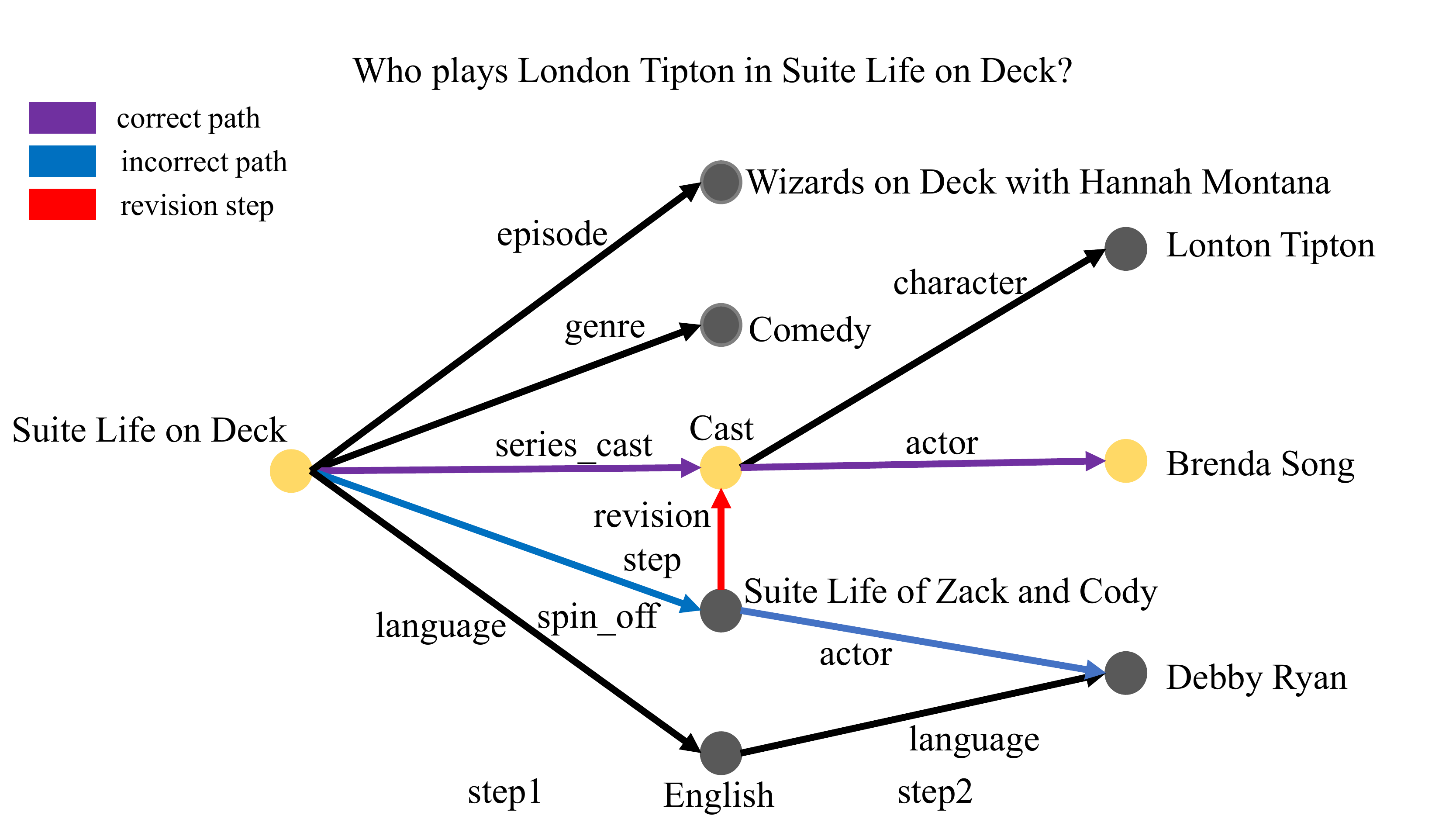}
\caption{One Case from WebQSP}
\label{case 2}
\end{minipage}
\hfill
\begin{minipage}[h]{0.5\linewidth}
\includegraphics[width=\linewidth]{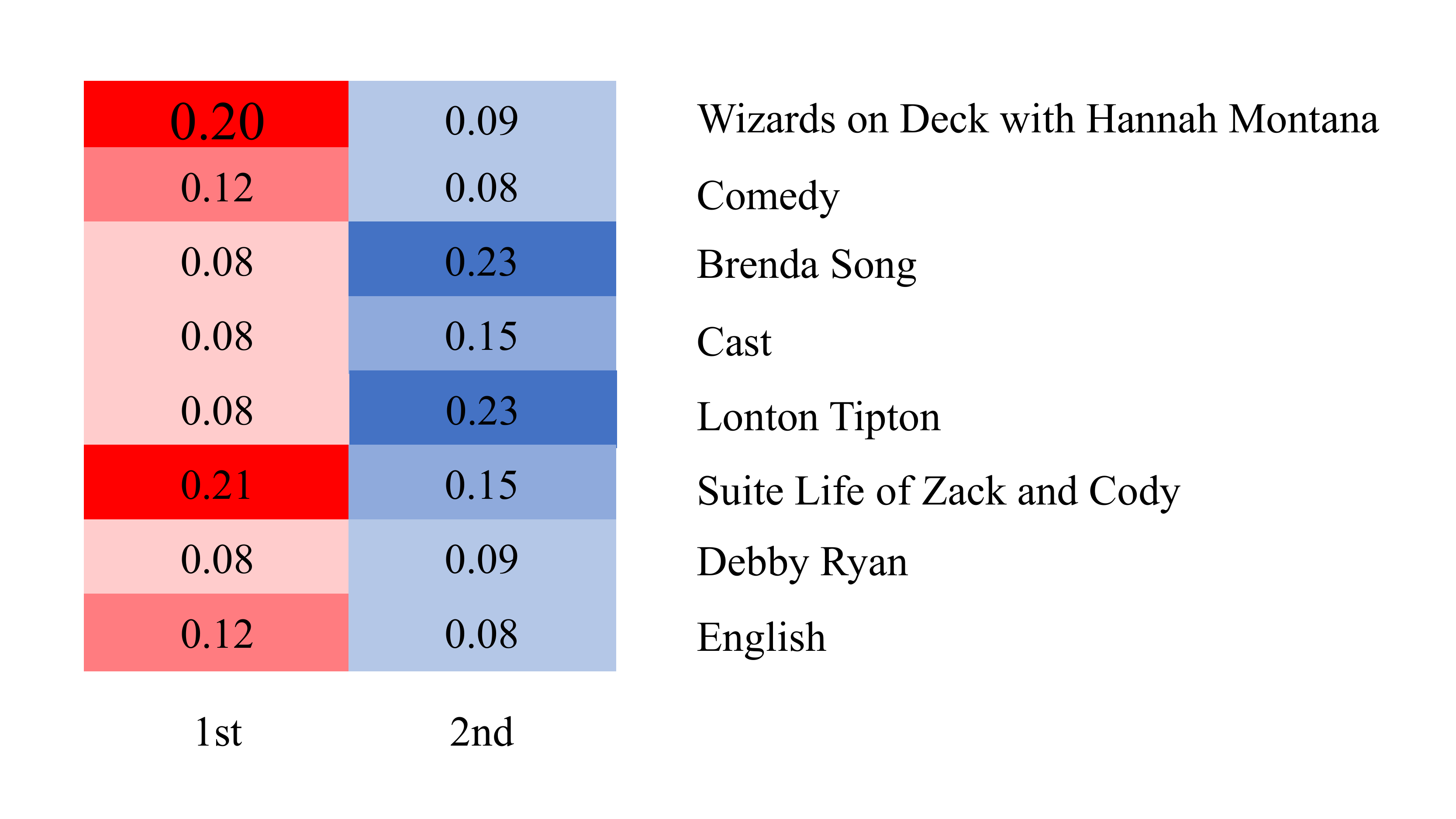}
\caption{The attention weights of entities in graph aggregation section}
\label{attention weights}
\end{minipage}
\vspace{-2em}
\end{figure}
In Figure \ref{case 2}, for the question ``Who plays London Tipton in Suite Life on Deck?'', the purple path denotes the right reasoning path: ``Suite Life on Deck (TV series)'' $\Longrightarrow{}_{series\_cast}$    ``cast'' $\Longrightarrow{}_{actor\_starring}$ ``Brenda Song (actor)'', where the entity ``cast'' satisfies the character constraint ``Lonton Tipton''. In the first step, our model and NSM predict the wrong intermediate entity ``Suite Life of Zack and Cody'' because of the high similarity between ``Suite Life on Deck'' and ``Suite Life of Zack and Cody''.
NSM fails to revise the wrong prediction of the intermediate entity for lack of graph information and derives the wrong answer along the blue path.
% considering ``Suite Life on Deck'' spins off from ``Suite Life of Zack and Cody''. 
By contrast, utilizing the knowledge graph information that ``Suite Life of Zack and Cody'' does not connect to any entity that satisfies the character constraint ``Lonton Tipton'', our model revises the error of incorrect prediction for intermediate entity and obtain the correct answer ``Brenda Song''. At the first step, the graph aggregation in our model attends to content-related entities such as ``Suite Life of Zack and Cody'', which are one hop away from the topic entity. At the second step, the attention weights of ``Brenda Song'' and ``Lonton Tipton'' rise from 0.08 to 0.23, indicating the graph structure around the answer entity is focused on by our model. This case shows the effectiveness of introducing knowledge graph into instructions generation on revising the incorrect prediction of an intermediate entity. 

\section{Conclusion and Future Work}
In this paper, we propose a novel and efficient approach KBIGER with the framework of iterative instruction generation and reasoning over the graph. We introduce the knowledge graph structure into instruction generation from the query question and it can revise the error of incorrect prediction for intermediate entities within the reasoning path. We conduct experiments on two benchmark datasets of this field and our approach outperforms all the existing methods. In the future, we will incorporate knowledge graph embedding into our framework to fuse the information from the query question and knowledge graph in a better manner.
 \bibliographystyle{splncs04}
 \bibliography{mybibliography}

\begin{thebibliography}{10}
\providecommand{\url}[1]{\texttt{#1}}
\providecommand{\urlprefix}{URL }
\providecommand{\doi}[1]{https://doi.org/#1}

\bibitem{berant2013semantic}
Berant, J., Chou, A., Frostig, R., Liang, P.: Semantic parsing on freebase from
  question-answer pairs. In: EMNLP. pp. 1533--1544 (2013)

\bibitem{bollacker2008freebase}
Bollacker, K., Evans, C., Paritosh, P., Sturge, T., Taylor, J.: Freebase: a
  collaboratively created graph database for structuring human knowledge. In:
  SIGMOD. pp. 1247--1250 (2008)

\bibitem{cohen2020scalable}
Cohen, W.W., Sun, H., Hofer, R.A., Siegler, M.: Scalable neural methods for
  reasoning with a symbolic knowledge base. ICLR  (2020)

\bibitem{do2019compact}
Do, T., Do, T.T., Tran, H., Tjiputra, E., Tran, Q.D.: Compact trilinear
  interaction for visual question answering. In: ICCV. pp. 392--401 (2019)

\bibitem{dong2015question}
Dong, L., Wei, F., Zhou, M., Xu, K.: Question answering over freebase with
  multi-column convolutional neural networks. In: ACL. pp. 260--269 (2015)

\bibitem{ferrucci2010building}
Ferrucci, D., Brown, E., Chu-Carroll, J., Fan, J., Gondek, D., Kalyanpur, A.A.,
  Lally, A., Murdock, J.W., Nyberg, E., Prager, J., et~al.: Building watson: An
  overview of the deepqa project. AI magazine  \textbf{31}(3),  59--79 (2010)

\bibitem{furlanello2018born}
Furlanello, T., Lipton, Z., Tschannen, M., Itti, L., Anandkumar, A.: Born again
  neural networks. In: ICML. pp. 1607--1616. PMLR (2018)

\bibitem{guo2018dialog}
Guo, D., Tang, D., Duan, N., Zhou, M., Yin, J.: Dialog-to-action:
  Conversational question answering over a large-scale knowledge base. In:
  NIPS. pp. 2942--2951 (2018)

\bibitem{he2021improving}
He, G., Lan, Y., Jiang, J., Zhao, W.X., Wen, J.R.: Improving multi-hop
  knowledge base question answering by learning intermediate supervision
  signals. In: WSDM. pp. 553--561 (2021)

\bibitem{hinton2015distilling}
Hinton, G., Vinyals, O., Dean, J.: Distilling the knowledge in a neural
  network. NIPS  (2015)

\bibitem{hu2018attention}
Hu, M., Peng, Y., Wei, F., Huang, Z., Li, D., Yang, N., Zhou, M.:
  Attention-guided answer distillation for machine reading comprehension. EMNLP
   (2018)

\bibitem{liang2016neural}
Liang, C., Berant, J., Le, Q., Forbus, K.D., Lao, N.: Neural symbolic machines:
  Learning semantic parsers on freebase with weak supervision. NIPS  (2016)

\bibitem{miller2016key}
Miller, A., Fisch, A., Dodge, J., Karimi, A.H., Bordes, A., Weston, J.:
  Key-value memory networks for directly reading documents. EMNLP  (2016)

\bibitem{pennington2014glove}
Pennington, J., Socher, R., Manning, C.D.: Glove: Global vectors for word
  representation. In: EMNLP. pp. 1532--1543 (2014)

\bibitem{qiu2020stepwise}
Qiu, Y., Wang, Y., Jin, X., Zhang, K.: Stepwise reasoning for multi-relation
  question answering over knowledge graph with weak supervision. In: WSDM. pp.
  474--482 (2020)

\bibitem{saha2019complex}
Saha, A., Ansari, G.A., Laddha, A., Sankaranarayanan, K., Chakrabarti, S.:
  Complex program induction for querying knowledge bases in the absence of gold
  programs. TACL  \textbf{7},  185--200 (2019)

\bibitem{saxena2020improving}
Saxena, A., Tripathi, A., Talukdar, P.: Improving multi-hop question answering
  over knowledge graphs using knowledge base embeddings. In: ACL. pp.
  4498--4507 (2020)

\bibitem{shi2021transfernet}
Shi, J., Cao, S., Hou, L., Li, J., Zhang, H.: Transfernet: An effective and
  transparent framework for multi-hop question answering over relation graph.
  EMNLP  (2021)

\bibitem{sun2019pullnet}
Sun, H., Bedrax-Weiss, T., Cohen, W.W.: Pullnet: Open domain question answering
  with iterative retrieval on knowledge bases and text. EMNLP  (2019)

\bibitem{sun2018open}
Sun, H., Dhingra, B., Zaheer, M., Mazaitis, K., Salakhutdinov, R., Cohen, W.W.:
  Open domain question answering using early fusion of knowledge bases and
  text. EMNLP  (2018)

\bibitem{talmor2018web}
Talmor, A., Berant, J.: The web as a knowledge-base for answering complex
  questions. NAACL  (2018)

\bibitem{yih2015semantic}
Yih, S.W.t., Chang, M.W., He, X., Gao, J.: Semantic parsing via staged query
  graph generation: Question answering with knowledge base. ACL  (2015)

\bibitem{zhang2018deep}
Zhang, Y., Xiang, T., Hospedales, T.M., Lu, H.: Deep mutual learning. In:
  Proceedings of the IEEE Conference on Computer Vision and Pattern
  Recognition. pp. 4320--4328 (2018)

\bibitem{zhang2018variational}
Zhang, Y., Dai, H., Kozareva, Z., Smola, A.J., Song, L.: Variational reasoning
  for question answering with knowledge graph. In: AAAI (2018)

\bibitem{zhao2019simple}
Zhao, W., Chung, T., Goyal, A., Metallinou, A.: Simple question answering with
  subgraph ranking and joint-scoring. NAACL  (2019)

\bibitem{zhou2018interpretable}
Zhou, M., Huang, M., Zhu, X.: An interpretable reasoning network for
  multi-relation question answering. COLING  (2018)

\end{thebibliography}
%





% LNCS Homepage, \url{http://www.springer.com/lncs}. Last accessed 4
%\end{thebibliography}
\end{document}